# Self-supervised Multi-level Face Model Learning
# for Monocular Reconstruction at over 250 Hz

Ayush Tewari[1,2]　Michael Zollhöfer[1,2,3]　Pablo Garrido[1,2]　Florian Bernard[1,2]
Hyeongwoo Kim[1,2]　Patrick Pérez[4]　Christian Theobalt[1,2]
[1]MPI Informatics　[2]Saarland Informatics Campus　[3]Stanford University　[4]Technicolor

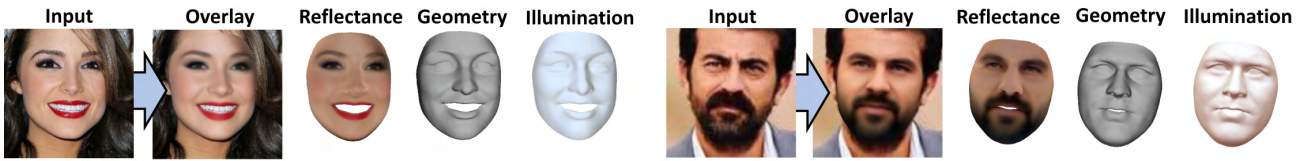

Our novel monocular reconstruction approach estimates high-quality facial geometry, skin reflectance (including facial hair) and incident illumination at over 250 Hz. A trainable multi-level face representation is learned jointly with the feed forward inverse rendering network. End-to-end training is based on a self-supervised loss that requires no dense ground truth.

## Abstract

*The reconstruction of dense 3D models of face geometry and appearance from a single image is highly challenging and ill-posed. To constrain the problem, many approaches rely on strong priors, such as parametric face models learned from limited 3D scan data. However, prior models restrict generalization of the true diversity in facial geometry, skin reflectance and illumination. To alleviate this problem, we present the first approach that jointly learns 1) a regressor for face shape, expression, reflectance and illumination on the basis of 2) a concurrently learned parametric face model. Our multi-level face model combines the advantage of 3D Morphable Models for regularization with the out-of-space generalization of a learned corrective space. We train end-to-end on in-the-wild images without dense annotations by fusing a convolutional encoder with a differentiable expert-designed renderer and a self-supervised training loss, both defined at multiple detail levels. Our approach compares favorably to the state-of-the-art in terms of reconstruction quality, better generalizes to real world faces, and runs at over 250 Hz.*

## 1. Introduction

Monocular face reconstruction has drawn an incredible amount of attention in computer vision and graphics in the last decades. The goal is to estimate a high-quality personalized model of a human face from just a single photograph. Such a model ideally comprises several interpretable semantic dimensions, e.g., 3D facial shape and expressions as well as surface reflectance properties. Research in this area is motivated by the increasing availability of face images, e.g., captured by webcams at home, as well as a wide range of important applications across several fields, such as facial motion capture, content creation for games and movies, virtual and augmented reality, and communication.

The reconstruction of faces from a single photograph is a highly challenging and ill-posed inverse problem, since the image formation process convolves multiple complex physical dimensions (geometry, reflectance and illumination) into a single color measurement per pixel. To deal with this ill-posedness, researchers have made additional prior assumptions, such as constraining faces to lie in a low-dimensional subspace, e.g., 3D Morphable Models (3DMM) [8] learned from scan databases of limited size. Many state-of-the-art optimization-based [6, 7, 52, 61, 26] and learning-based face reconstruction approaches [16, 48, 49, 62, 60] heavily rely on such priors. While these algorithms yield impressive results, they do not generalize well beyond the restricted low-dimensional subspace of the underlying model. Consequently, the reconstructed 3D face may lack important facial details, contain incorrect facial features and not align well to an image. For example, beards have shown to drastically deteriorate the reconstruction quality of algorithms that are trained on pure synthetic data [48, 49, 54] or employ a 3DMM for regularization [8, 61, 26, 62, 60]. Some approaches try to prevent these failures via heuristics, e.g., a separate segmentation method to disambiguate disjunct skin and hair regions [52]. Recent methods refine a fitted prior by adding fine-scale details, either based on shape-from-shading [26, 48] or pre-learned regressors [16, 49]. However, these approaches rely on slow optimization or require a high-quality annotated training corpus. Besides, they do not build an improved subspace of medium-scale shape, reflectance and expression, which is critical for generalization. Very recently, Sela et al. [54] predicted a per-pixel depth map to deform and fill holes of a limited geometry subspace learned during training. While the results are impressive, the non-rigid registration runs offline. Furthermore, their method captures face geometry only and fails if the faces differ drastically from the training corpus, e.g., in terms of skin reflectance and facial hair. Ideally, one would like to build better priors that explain a rich variety of real-world faces with meaningful and interpretable parameters. Learning such models in the traditional way requires large amounts of densely labeled real world data, which is practically infeasible.

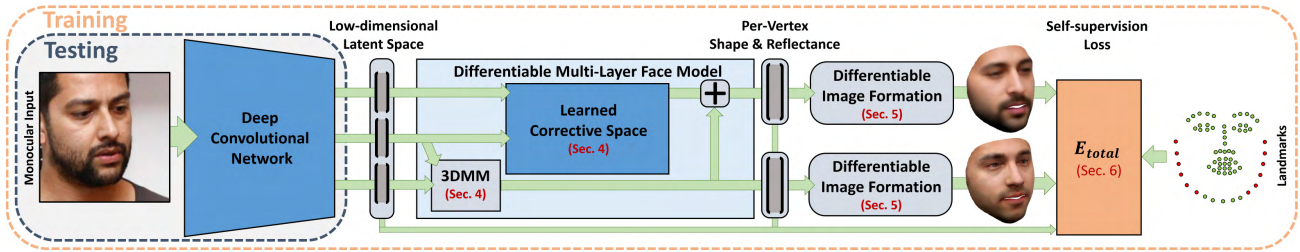

Figure 1. Our approach regresses a low-dimensional latent face representation at over 250 Hz. The feed forward CNN is jointly learned with a multi-level face model that goes beyond the low-dimensional subspace of current 3DMMs. Trainable layers are shown in blue and expert-designed layers in gray. Training is based on differentiable image formation in combination with a self-supervision loss (orange).

We present an entirely new end-to-end trainable method that jointly learns 1) an efficient regressor to estimate high-quality identity geometry, face expression, and colored skin reflectance, alongside 2) the parameterization of an improved multi-level face model that better generalizes and explains real world face diversity. Our method can be trained end-to-end on sparsely labeled in-the-wild images and reconstructs face and illumination from monocular RGB input at over 250 Hz. Our approach takes advantage of a 3DMM for regularization and a learned corrective space for out-of-space generalization. To make end-to-end training on in-the-wild images feasible, we propose a hybrid convolutional auto-encoder that combines a CNN encoder with a differentiable expert-designed rendering layer and a self-supervision loss, both defined at multiple levels of details. In addition, we incorporate a novel contour constraint that generates a better face alignment. Unlike Tewari et al. [60], our auto-encoder learns an improved multi-level model that goes beyond a predefined low-dimensional parametric face prior. Experimental evaluations show that our approach is more robust, generalizes better, and estimates geometry, reflectance and lighting at higher quality.

## 2. Related Work

We focus our discussion on optimization- and learning-based approaches that employ parametric models. While high-quality multi-view 3D reconstruction methods [4, 13, 5, 24, 30, 69] exist, we are interested in the harder monocular reconstruction problem.

**Parametric Face Models**: The most widely used face model is the 3D Morphable Model (3DMM) [8], which is an affine parametric model of face geometry and texture that is learned from high quality scans. A similar model for facial animation is presented in [6]. Recently, Booth et al. [11] created a Large-scale Facial Model (LSFM) from around 10,000 facial scans which represents a richer shape distribution. In Booth et al. [10], the face model is augmented with an 'in-the-wild' texture model. Fitting such a model to an image is a non-convex optimization problem, akin to frameworks based on Active Shape (ASMs) [22] and Appearance (AAMs) [21] Models. Although 3DMMs are highly efficient priors, they limit face reconstruction to a restricted low-dimensional subspace, e.g., beards or characteristic noses can not be reconstructed. We, on the contrary, extend the limited subspace by jointly learning a correction model that generalizes much better to real-world data.

**Optimization-based Approaches**: Many approaches for monocular face reconstruction [50], reconstruction based on image collections [51], and the estimation of high-quality 3D face rigs [26] are based on energy optimization. Impressive face reconstruction results have been obtained from varying data sources, e.g., photo collections [36], internet photos [35] or videos [59]. Also, methods that do not rely on a trained shape or appearance model have been proposed, e.g., they use a model obtained using modal analysis [1], or leverage optical flow in combination with message-passing [25]. While real-time face tracking is in general feasible [61, 31], optimization-based face reconstruction is computationally expensive. Moreover, optimization-based approaches are sensitive to initialization, thus requiring 2D landmark detection [66, 34]. Some approaches allow the 3D face silhouette to slide over a predefined path (e.g., isolines) [17, 71] or iterate over a fixed vertex set to find 3D contour correspondences [23]. Our approach requires neither an expensive optimization strategy nor parameter initialization, yet it accurately fits a 3D face mesh to an image by taking silhouettes into account during training.

**Learning-based Approaches**: In addition to optimization-based reconstruction approaches, there are many learning-based methods [72, 41, 27, 20, 56]. Among them there are methods that learn to detect fiducial points in images with high accuracy, e.g., based on CNNs [58, 70, 15] or Restricted Boltzmann Machines [67]. Furthermore, we can also find (weakly) supervised deep networks that integrate generative models to solve tasks like facial performance capture [39]. Ranja et al. [47] proposed a multi-purpose CNN for regressing semantic parameters (e.g., age, gender, pose) from face images. Richardson et al. [48] proposed a hybrid learning- and optimization-based method that reconstructs detailed facial geometry from a single image. The work presented in [49] train an end-to-end regressor to recover facial geometry at a coarse- and fine-scale level. In [62], face shape and texture are regressed for face identification. The generalization of the latter face reconstruction approaches ([48, 49, 62]) to the real-world diversity of face is limited by the underlying low-dimensional face model.

**Corrective Basis and Subspace Learning**: Face reconstruction quality can be improved by adding medium-scale detail. Li *et al.* [40] use incremental PCA for on-the-fly personalization of the expression basis. Bouaziz *et al.* [12] introduced medium-scale shape correctives based on manifold harmonics [65]. Recently, Garrido *et al.* [26] proposed to learn medium-scale shape from a monocular video based on a fixed corrective basis. Sela et al. [54] directly regress depth and per-pixel correspondence, thus going beyond the restricted subspace of a 3DMM. Nonetheless, they do not recover colored surface reflectance and require an off-line non-rigid registration step to obtain reconstructions with known consistent topology. To the best of our knowledge, there is no algorithm that jointly learns geometry and reflectance correctives from in-the-wild images.

**Deep Integration of Generative Models**: The seminal work by Jaderberg *et al.* [32] introduced *spatial transformer nets* that achieve pose-invariance within a neural network. The authors of [3] extend this work by using a 3DMM as spatial transformer network. *Perspective transformer nets* [68] are able to obtain a 3D object representation from a single 2D image. The *gvvn* library [28] implements low-level computer vision layers for such transformations. Recently, a model-based face autoencoder (MoFA) [60] has been proposed for monocular face reconstruction that combines an expert-designed rendering layer with a trainable CNN encoder. Their results are remarkable but limited to the fixed low-dimensional subspace of the face model. Out-of-subspace variation, e.g., facial detail and personalized noses, are not reproduced and severely degrades the reconstruction quality. Our approach addresses all these challenges, achieving more robustness and higher quality in terms of geometry and reflectance.

## 3. Method Overview

Our novel face reconstruction approach estimates high-quality geometry, skin reflectance and incident illumination from a single image. We jointly train a regressor for all dimensions on the basis of a concurrently learned multi-level parametric face model, see Fig. 1.

**Parameter Regression**: At test time (Fig. 1, left), a low-dimensional, yet expressive and discriminative, latent space face representation is computed in under 4ms using a feed forward CNN, e.g., AlexNet [38] or VGG-Face [45]. Our latent space is based on a novel multi-level face model (Sec. 4) that combines a coarse-scale 3DMM with trainable per-vertex geometry and skin reflectance correctives. This enables our approach to go beyond the restricted low-dimensional geometry and skin reflectance subspaces, commonly used by 3DMM-based methods for face fitting.

**Self-Supervised Training**: We train (Fig. 1, right) the feed forward network jointly with the corrective space based on a novel CNN architecture that does not rely on a densely annotated training corpus of ground truth geometry, skin reflectance and illumination. To this end, we combine the multi-level model with an expert-designed image formation layer (Sec. 5) to obtain a differentiable computer graphics module. To enable the joint estimation of our multi-level face model, this module renders both the coarse 3DMM model and the medium-scale model that includes the correctives. For training, we employ self-supervised loss functions (Sec. 6) to enable efficient end-to-end training of our architecture on a large corpus of in-the-wild face images without the need for densely annotated ground truth. We evaluate our approach qualitatively and quantitatively, and compare it to state-of-the-art optimization and learning-based face reconstruction techniques (see Sec. 7).

## 4. Trainable Multi-level Face Model

At the core of our approach is a novel multi-level face model that parameterizes facial geometry and skin reflectance. Our model is based on a manifold template mesh with $N \sim 30k$ vertices and per-vertex skin reflectance. We stack the $x$-, $y$- and $z$-coordinates of all vertices $\mathbf{v}_i \in \mathcal{V}$ in a geometry vector $\mathbf{v}^f \in \mathbb{R}^{3N}$. Similarly, we obtain a vector of per-vertex skin reflectance $\mathbf{r}^f \in \mathbb{R}^{3N}$. We parameterize geometry and reflectance as follows:

$$\mathbf{v}^f(\mathbf{x}_g) = \mathbf{v}^b(\boldsymbol{\alpha}) + \mathcal{F}_g(\boldsymbol{\delta}_g|\Theta_g) \in \mathbb{R}^{3N} \text{(geometry)}, \quad (1)$$

$$\mathbf{r}^f(\mathbf{x}_r) = \mathbf{r}^b(\boldsymbol{\beta}) + \mathcal{F}_r(\boldsymbol{\delta}_r|\Theta_r) \in \mathbb{R}^{3N} \text{(reflectance)}, \quad (2)$$

where $\mathbf{x}_g = (\boldsymbol{\alpha}, \boldsymbol{\delta}_g, \Theta_g)$ and $\mathbf{x}_r = (\boldsymbol{\beta}, \boldsymbol{\delta}_r, \Theta_r)$ are the geometry and reflectance parameters, respectively. At the base level is an affine face model that parameterizes the (coarse) facial geometry $\mathbf{v}^b$ and (coarse) skin reflectance $\mathbf{r}^b$ via a low-dimensional set of parameters $(\boldsymbol{\alpha}, \boldsymbol{\beta})$. In addition, we employ correctives to add medium-scale geometry $\mathcal{F}_g$ and reflectance $\mathcal{F}_r$ deformations, parametrized by $(\boldsymbol{\delta}_g, \Theta_g)$ and $(\boldsymbol{\delta}_r, \Theta_r)$, respectively. A detailed explanation will follow in Sec. 4.2. A combination of the base level model with the corrective model yields the final level model, parameterizing $\mathbf{v}^\text{f}$ and $\mathbf{r}^\text{f}$. In the following, we describe the different levels of our multi-level face model.

### 4.1. Static Parametric Base Model

The parametric face model employed on the base level expresses the space of plausible facial geometry and reflectance via two individual affine models:

$$\mathbf{v}^\text{b}(\boldsymbol{\alpha}) = \mathbf{a}_g + \sum_{k=1}^{m_s+m_e} \boldsymbol{\alpha}_k \mathbf{b}_k^g \quad \text{(geometry)}, \quad (3)$$

$$\mathbf{r}^\text{b}(\boldsymbol{\beta}) = \mathbf{a}_r + \sum_{k=1}^{m_r} \boldsymbol{\beta}_k \mathbf{b}_k^r \quad \text{(reflectance)}. \quad (4)$$

Here, $\mathbf{a}_g \in \mathbb{R}^{3N}$ is the average facial geometry and $\mathbf{a}_r \in \mathbb{R}^{3N}$ the corresponding average reflectance. The subspace of reflectance variations is spanned by the vectors $\{\mathbf{b}_k^r\}_{k=1}^{m_r}$, created using PCA from a dataset of 200 high-quality face scans (100 male, 100 female) of Caucasians [8]. The geometry subspace is split into $m_s$ and $m_e$ modes, representing shape and expression variations, respectively. The vectors spanning the subspace of shape variations $\{\mathbf{b}_k^g\}_{k=1}^{m_s}$ are constructed from the same data as the reflectance space [8]. The subspace of expression variations is spanned by the vectors $\{\mathbf{b}_k^g\}_{k=m_s+1}^{m_s+m_e}$ which were created using PCA from a subset of blendshapes of [2] and [18]. Note that these blendshapes have been transferred to our topology using deformation transfer [57]. The basis captures 99% of the variance of the used blendshapes. We employ $m_s = m_r = 80$ shape and reflectance vectors, and $m_e = 64$ expression vectors. The associated standard deviations $\boldsymbol{\sigma}_g$ and $\boldsymbol{\sigma}_r$ have been computed assuming a normally distributed population. The model parameters $(\boldsymbol{\alpha}, \boldsymbol{\beta}) \in \mathbb{R}^{80+64} \times \mathbb{R}^{80}$ constitute a low-dimensional encoding of a particular face. Even though such a parametric model provides a powerful prior, its low dimensionality is a severe weakness as it can only represent coarse-scale geometry.

### 4.2. Trainable Shape and Reflectance Corrections

Having only a coarse-scale face representation is one of the major shortcomings of many optimization- and learning-based reconstruction techniques, such as [8, 6, 61, 60]. Due to its low dimensionality, the base model described in Sec. 4.1 has a limited expressivity for modeling the facial shape and reflectance at high accuracy. A particular problem is skin albedo variation, since the employed model has an ethnic bias and lacks facial hair, e.g., beards. The purpose of this work is to improve upon this by learning a trainable corrective model that can represent these out-of-space variations. Unlike other approaches that use a fixed pre-defined corrective basis [26], we learn both the generative model for correctives and the best corrective parameters. Furthermore, we require no ground truth annotations for geometry, skin reflectance and incident illumination.

Our corrective model is based on (potentially non-linear) mappings $\mathcal{F}_\bullet : \mathbb{R}^C \to \mathbb{R}^{3N}$ that map the $C$-dimensional corrective parameter space onto per-vertex corrections in shape or reflectance. The mapping $\mathcal{F}_\bullet(\delta_\bullet|\Theta_\bullet)$ is a function of $\delta_\bullet \in \mathbb{R}^C$ that is parameterized by $\Theta_\bullet$. The motivation for disambiguating between $\delta_\bullet$ and $\Theta_\bullet$ is that during training we learn both $\delta_\bullet$ and $\Theta_\bullet$, while at test time we keep $\Theta_\bullet$ fixed and directly regress the corrective parameters $\delta_\bullet$ using the feed forward network. In the affine/linear case, one can interpret $\Theta_\bullet$ as a basis that spans a subspace of the variations, and $\delta_\bullet$ is the coefficient vector that reconstructs a given sample using the basis. However, in general we do not assume $\mathcal{F}_\bullet$ to be affine/linear. The key difference to the base level is that the correction level does not use a fixed pre-trained basis but learns a generative model, along with the coefficients, directly from the training data.

## 5. Differentiable Image Formation Model

To train our novel multi-level face reconstruction approach end-to-end, we require a differentiable image formation model. In the following, we describe its components.

**Full Perspective Camera**: We parameterize the position and rotation of the virtual camera based on a rigid transformation $\Phi(\mathbf{v}) = \mathbf{R}\mathbf{v} + \mathbf{t}$, which maps a model space 3D point $\mathbf{v}$ onto camera space $\hat{\mathbf{v}} = \Phi(\mathbf{v})$. Here, $\mathbf{R} \in \text{SO}(3)$ is the camera rotation and $\mathbf{t} \in \mathbb{R}^3$ is the translation vector. To render virtual images of the scene, we use a full perspective camera model to project the camera space point $\hat{\mathbf{v}}$ into a 2D point $\mathbf{p} = \Pi(\hat{\mathbf{v}}) \in \mathbb{R}^2$. The camera model contains the intrinsics and performs the perspective division.

**Illumination Model**: We make the assumption of distant lighting and approximate the incoming radiance using spherical harmonics (SH) basis functions $H_b : \mathbb{R}^3 \to \mathbb{R}$. We assume that the incoming radiance only depends on the surface normal $\mathbf{n}$:

$$\tilde{B}(\mathbf{r}, \mathbf{n}, \boldsymbol{\gamma}) = \mathbf{r} \odot \sum_{b=1}^{B^2} \boldsymbol{\gamma}_b H_b(\mathbf{n}) \ . \quad (5)$$

Here, $\odot$ denotes the Hadamard product, $\mathbf{r}$ is the surface reflectance and $B$ is the number of spherical harmonics bands. $\boldsymbol{\gamma}_b \in \mathbb{R}^3$ are coefficients to control the illumination. Since the incident radiance is sufficiently smooth, an average error below 1% [46] can be achieved with only $B = 3$ bands independent of the illumination. This leads to $m_l = B^2 = 9$ variables per color channel.

**Image Formation**: Our differentiable image formation layer takes as input the per-vertex shape and reflectance in model space. This can be the model of the base level $\mathbf{v}^b$ and $\mathbf{r}^b$ or of the final level $\mathbf{v}^f$ and $\mathbf{r}^f$ that include the learned correctives. Let $\mathbf{v}_i^\ell \in \mathbb{R}^3$ and $\mathbf{r}_i^\ell \in \mathbb{R}^3$ denote the position and the reflectance of the $i$-th vertex for the base level ($\ell$ = b) and the final level ($\ell$ = f). Our rendering layer takes this information and forms a point-based rendering of the scene, as follows. First, it maps the points onto camera space, i.e., $\hat{\mathbf{v}}_i^\ell = \Phi(\mathbf{v}_i^\ell)$, and then computes the projected pixel positions of all vertices as

$$\mathbf{u}_i^\ell(\mathbf{x}) = \Pi(\hat{\mathbf{v}}_i^\ell) \ .$$

The shaded colors $\mathbf{c}_i^\ell$ at these pixel locations are computed based on the illumination model described before:

$$\mathbf{c}_i^\ell(\mathbf{x}) = \tilde{B}(\mathbf{r}_i^\ell, \hat{\mathbf{n}}_i^\ell, \boldsymbol{\gamma}) \ ,$$

where $\hat{\mathbf{n}}_i^\ell$ are the associated camera space normals to $\hat{\mathbf{v}}_i^\ell$. Our image formation model is differentiable, which enables end-to-end training using back propagation. The free variables that the regressor learns to predict are: The model parameters $(\boldsymbol{\alpha}, \boldsymbol{\beta}, \boldsymbol{\delta}_g, \boldsymbol{\delta}_r)$, the camera parameters $\mathbf{R}, \mathbf{t}$ and the illumination parameters $\boldsymbol{\gamma}$. In addition, during training, we learn the corrective shape and reflectance bases $\Theta_g, \Theta_r$. This leads to the following vector of unknowns:

$$\mathbf{x} = (\boldsymbol{\alpha}, \boldsymbol{\beta}, \boldsymbol{\delta}_g, \boldsymbol{\delta}_r, \mathbf{R}, \mathbf{t}, \boldsymbol{\gamma}, \Theta_g, \Theta_r) \in \mathbb{R}^{257+2C+|\Theta_g|+|\Theta_r|} \ .$$

## 6. Self-supervised Learning

Our face regression network is trained using a novel self-supervision loss that enables us to fit our base model and learn per-vertex correctives end-to-end. Our loss function consists of a data fitting and regularization term:

$$E_{\text{total}}(\mathbf{x}) = E_{\text{data}}(\mathbf{x}) + w_{\text{reg}} E_{\text{reg}}(\mathbf{x}) \ , \quad (6)$$

where $E_{\text{data}}$ penalizes misalignments of the model to the input image and $E_{\text{reg}}$ encodes prior assumptions about faces at the coarse and medium scale. Here, $w_{\text{reg}}$ is a trade-off factor that controls the amount of regularization. The data fitting term is based on sparse and dense alignment constraints:

$$E_{\text{data}}(\mathbf{x}) = E_{\text{sparse}}(\mathbf{x}) + w_{\text{photo}} E_{\text{photo}}(\mathbf{x}) \ . \quad (7)$$

The regularization term represents prior assumptions on the base and corrective model:

$$E_{\text{reg}}(\mathbf{x}) = E_{\text{std}}(\mathbf{x}) + E_{\text{smo}} + E_{\text{ref}}(\mathbf{x}) + E_{\text{glo}}(\mathbf{x}) + E_{\text{sta}}(\mathbf{x}) \ . \quad (8)$$

In the following, the individual terms are explained in detail.

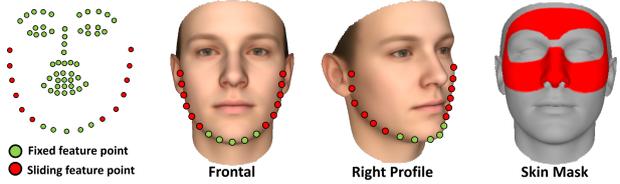

Figure 2. We distinguish between fixed and sliding feature points. This leads to better contour alignment. Note how the outer contour depends on the rigid head pose (left). The skin mask (right) is employed in the global reflectance constancy constraint.

### 6.1. Data Terms

**Multi-level Dense Photometric Loss**: We employ a dense multi-level photometric loss function that measures the misalignment of the coarse and fine fit to the input. Let $\bar{\mathcal{V}}$ be the set of all visible vertices. Our photometric term is then defined as:

$$E_{\text{photo}}(\mathbf{x}) = \sum_{\ell \in \{b,f\}} \frac{1}{N} \sum_{i \in \bar{\mathcal{V}}} \left\| \mathcal{I}\big(\mathbf{u}_i^\ell(\mathbf{x})\big) - \mathbf{c}_i^\ell(\mathbf{x}) \right\|_2 \ . \quad (9)$$

Here, $\mathbf{u}_i^\ell(\mathbf{x})$ is the screen space position, $\mathbf{c}_i^\ell(\mathbf{x})$ is the shaded color of the $i$-th vertex, and $\mathcal{I}$ is the current image during training. For robustness, we employ the $\ell_{2,1}$-norm, which measures the color distance using the $\ell_2$-norm, while the summation over all pixel-wise $\ell_2$-norms encourages sparsity as it corresponds to the $\ell_1$-norm. Visibility is computed using backface culling. This is an approximation, but works well, since faces are almost convex.

**Sparse Feature Points**: Faces contain many salient feature points. We exploit this by using a weak supervision in the form of automatically detected 66 facial landmarks $\mathbf{f} \in \mathcal{F} \subset \mathbb{R}^2$ [53] and associated confidence $c_\mathbf{f} \in [0,1]$ (1 confident). The set of facial landmarks falls in two categories: Fixed and sliding feature points. Fixed feature points, e.g. eyes and nose, are associated with a fixed vertex on the template model, whereas sliding feature points, e.g., the face contour, change their position on the template based on the rigid pose, see Fig. 2. We explicitly model this as follows:

$$E_{\text{sparse}}(\mathbf{x}) = \frac{1}{|\mathcal{F}|} \sum_{\mathbf{f} \in \mathcal{F}} c_\mathbf{f} \cdot \left\| \mathbf{f} - \mathbf{u}_{k_\mathbf{f}}^b(\mathbf{x}) \right\|_2^2 \ . \quad (10)$$

Here, $k_\mathbf{f}$ is the index of the target vertex. For fixed feature points, we hard-code the index of the corresponding mesh vertex. The indexes for sliding feature points are computed via an alternation scheme: In each step of stochastic gradient descent, we find the mesh vertex that is closest to the 3D line, defined by the camera center and the back-projection of the detected 2D feature point. Based on the squared Euclidean distance we set $k_\mathbf{f}$ to the index of the closest vertex.

## 6.2. Regularization Terms

**Statistical Regularization**: We enforce statistical regularization on the 3DMM model parameters of the base level to ensure plausible reconstructions. Based on the assumption that the model parameters follow a zero-mean Gaussian distribution, we employ Tikhonov regularization:

$$E_{\text{std}}(\mathbf{x}) = \sum_{k=1}^{m_s+m_e} \left(\frac{\alpha_k}{(\boldsymbol{\sigma}_g)_k}\right)^2 + w_{\text{rstd}} \sum_{k=1}^{m_r} \left(\frac{\beta_k}{(\boldsymbol{\sigma}_r)_k}\right)^2 \quad . \quad (11)$$

This is a common constraint [8, 61, 26, 60] that prevents the degeneration of the facial geometry and face reflectance in the ill-posed monocular reconstruction scenario.

**Corrective Smoothness**: We also impose local smoothness by adding Laplacian regularization on the vertex displacements for the set of all vertices $\mathcal{V}$:

$$E_{\text{smo}}(\mathbf{x}) = \frac{w_{\text{smo}}}{N} \sum_{i \in \mathcal{V}} \left\| \frac{1}{|\mathcal{N}_i|} \sum_{j \in \mathcal{N}_i} \left( (\mathcal{F}_g(\mathbf{x}))_i - (\mathcal{F}_g(\mathbf{x}))_j \right) \right\|_2^2. \quad (12)$$

Here, $(\mathcal{F}_g(\mathbf{x}))_i = (\mathcal{F}_g(\boldsymbol{\delta}_g | \Theta_g))_i$ denotes the correction for the $i$-th vertex given the parameter $\mathbf{x}$, and $\mathcal{N}_i$ is the 1-ring neighborhood of the $i$-th vertex.

**Local Reflectance Sparsity**: In spirit of recent intrinsic decomposition approaches [9, 43], we enforce sparsity to further regularize the reflectance of the full reconstruction:

$$E_{\text{ref}}(\mathbf{x}) = w_{\text{ref}} \frac{1}{N} \sum_{i \in \mathcal{V}} \sum_{j \in \mathcal{N}_i} w_{i,j} \cdot \left\| \mathbf{r}_i^{\text{f}}(\mathbf{x}) - \mathbf{r}_j^{\text{f}}(\mathbf{x}) \right\|_2^p \quad . \quad (13)$$

Here, $w_{i,j} = \exp(-\alpha \cdot \|\mathcal{I}(u_i^{\text{f}}(\mathbf{x}^{\text{old}})) - \mathcal{I}(u_j^{\text{f}}(\mathbf{x}^{\text{old}}))\|_2)$ are constant weights that measure the chromaticity similarity between the colors in the input, where $\mathbf{x}^{\text{old}}$ are the parameters estimated in the previous iteration. We assume that pixels with the same chromaticity are more likely to have the same reflectance. The term $\|\cdot\|_2^p$ enforces sparsity on the combined reflectance estimate. We employ $\alpha = 50$ and $p = 0.9$ in all our experiments.

**Global Reflectance Constancy**: We enforce skin reflectance constancy over a fixed set of vertices that covers only the skin region, see Fig. 2 (right):

$$E_{\text{glo}}(\mathbf{x}) = w_{\text{glo}} \frac{1}{|\mathcal{M}|} \sum_{i \in \mathcal{M}} \sum_{j \in \mathcal{G}_i} \left\| \mathbf{r}_i^{\text{f}}(\mathbf{x}) - \mathbf{r}_j^{\text{f}}(\mathbf{x}) \right\|_2^2 \quad . \quad (14)$$

Here, $\mathcal{M}$ is the per-vertex skin mask and $\mathcal{G}_i$ stores 6 random samples of vertex indexes of the mask region. The idea is to enforce the whole skin region to have the same reflectance. For efficiency, we use reflectance similarity between random pairs of vertices in the skin region. Note that regions that may have facial hair were not included in the mask. In combination, local and global reflectance constancy efficiently removes shading from the reflectance channel.

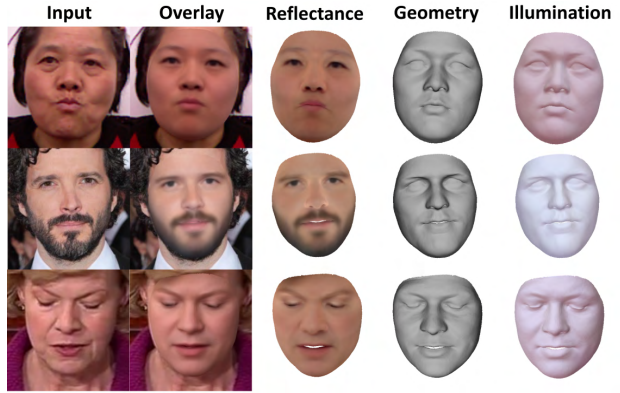

Figure 3. Our approach allows for high-quality reconstruction of facial geometry, reflectance and incident illumination from just a single monocular color image. Note the reconstructed facial hair, e.g., the beard, reconstructed make-up, and the eye lid closure, which are outside of the space of the used 3DMM.

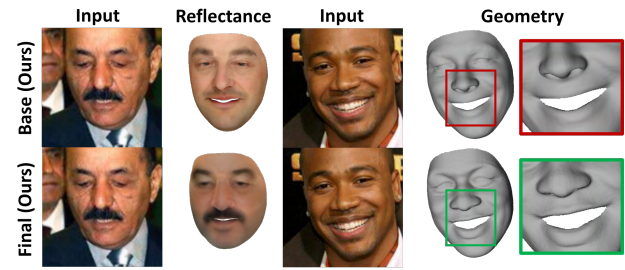

Figure 4. Jointly learning a multi-level model improves the geometry and reflectance compared to the 3DMM. Note the better aligning nose, lips and the reconstructed facial hair.

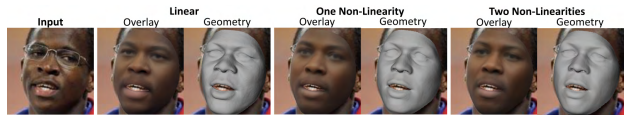

Figure 5. Comparison of linear and non-linear corrective spaces.

**Stabilization**: We also ensure that the corrected geometry stays close to the base reconstruction by enforcing small vertex displacements:

$$E_{\text{sta}}(\mathbf{x}) = w_{\text{sta}} \frac{1}{N} \sum_{i \in \mathcal{V}} \left\| (\mathcal{F}_g(\mathbf{x}))_i \right\|_2^2 \quad . \quad (15)$$

## 7. Results

We demonstrate joint end-to-end self-supervised training of the feed forward encoder and our novel multi-level face representation based on in-the-wild images without the need for densely annotated ground truth. Our approach regresses pose, shape, expression, reflectance and illumination at high-quality with over 250 Hz, see Fig. 3. For the feed forward encoder we employ a modified version of AlexNet [38] that outputs the parameters of our face model. Note that other feed forward architectures could be used. We implemented our approach using *Caffe* [33]. Training is based on *AdaDelta* with a batch size of 5. We pretrain our network up to the base level for 200k iterations with a learning rate of 0.01. Afterwards, we finetune our complete network for 190k iterations with a learning rate of 0.001 for the base level, 0.005 for the geometry and 0.01 for the reflectance correctives. All components of our network are implemented in CUDA [44] for efficient training, which takes 16 hours. We use the same weights $w_\bullet$ in all experiments. In the following, we fix the size, $C$ of the corrective parameters to 500 for both geometry and reflectance. We tested different corrective spaces (linear and non-linear), see Fig. 5. A linear corrective basis gave the best results, so we use it for all following experiments. Please refer to the supplemental document and video[1] for more details. Our approach is trained on a corpus of in-the-wild face images, without densely annotated ground truth. We combined four different datasets: CelebA [42], LFW [29], FaceWarehouse [17], and 300-VW [19, 55, 63]. Sparse landmark annotations are obtained automatically [53] and we crop to a tight face bounding box of $240 \times 240$ pixels using Haar Cascade Face Detection [14]. Images with bad detections are automatically removed based on landmark confidence. In total, we use 144k images, which we split into a training (142k images) and validation (2k images) set.

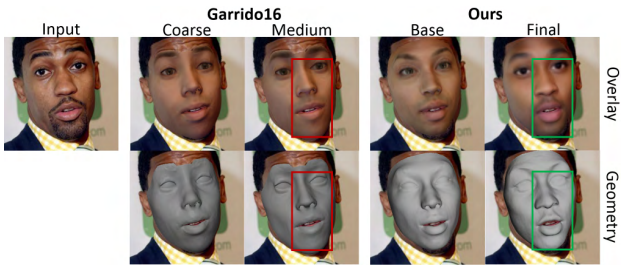

Figure 6. Comparison to Garrido et al. [26]. We achieve higher quality reconstructions, since our jointly learned model generalizes better than a corrective space based on manifold harmonics.

We compare our final output ('final') to the base low-dimensional 3DMM reconstruction ('base') obtained from the pretrained network to illustrate that our multi-level model lets us recover higher quality geometry and reflectance (Fig. 4). In the following, we show more results, evaluate our approach, and compare to the state-of-the-art.

## 7.1. Comparisons to the State-of-the-art

**Optimization-based Techniques**: We compare to the optimization-based high-quality reconstruction method of Garrido et al. [26], see Fig. 6. Our approach obtains similar geometry quality but better captures the person's characteristics due to our learned corrective space. Since our approach jointly learns a corrective reflectance space, it can leave the restricted subspace of the underlying 3DMM and thus produces more realistic appearance. Note, unlike Garrido et al., our approach does not require landmarks at test time and runs orders of magnitude faster (4ms vs. 120s per image). We also compare to the approach of Booth et al. [10], see Fig. 7. Our approach jointly learns a better shape and reflectance model, while their approach only builds an 'in-the-wild' texture model that contains shading. In contrast to our approach, Booth et al. is based on optimization and requires initialization or landmarks.

[1] http://gvv.mpi-inf.mpg.de/projects/FML

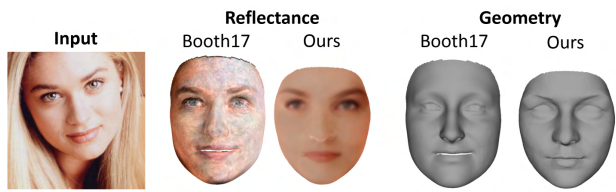

Figure 7. In contrast to the texture model of Booth et al. [10] that contains shading, our approach yields a reflectance model.

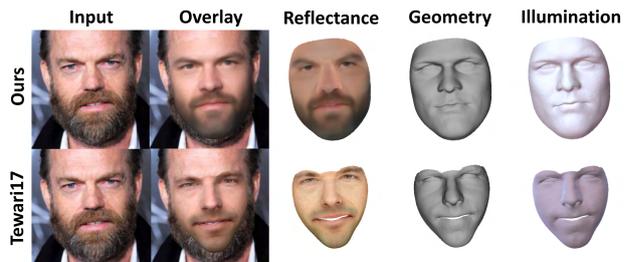

Figure 8. Comparison to [60]. We achieve higher quality (without surface shrinkage), due to our jointly trained model.

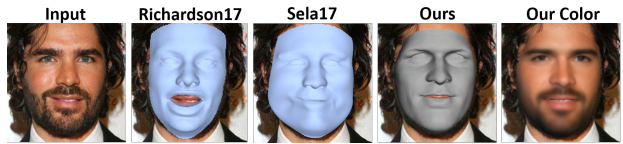

Figure 9. Comparison to [48, 49, 54]. They obtain impressive results within the span of the synthetic training corpus, but do not handle out-of-subspace variations, e.g., beards. Our approach is robust to hair and make-up, since the model is jointly learned.

**Learning-based Techniques**: We compare to the high-quality learning-based reconstruction approaches of Tewari et al. [60] (Fig. 8), Richardson et al. [48, 49] (Fig. 9) and Sela et al. [54] (Fig. 9). These approaches obtain impressive results within the span of the used synthetic training corpus or the employed 3DMM model, but suffer from out-of-subspace shape and reflectance variations, e.g., people with beards. Our approach is not only robust to facial hair and make-up, but also automatically learns to reconstruct such variations based on the jointly learned model. Reconstruction requires 4 ms, while [54] requires slow off-line non-rigid registration to obtain a hole free reconstruction from the predicted depth map. In addition, we jointly obtain a reconstruction of colored reflectance and illumination. Due to our model learning, our approach is able to leave the low-dimensional space of the 3DMM, which leads to a more realistic reconstruction of facial appearance and geometry.

Table 1. Geometric error on FaceWarehouse [17]. Our approach outperforms the deep learning techniques of [60] and [37]. It comes close to the high-quality approach of [26], while being orders of magnitude faster and not requiring feature detection.

| | **Ours** | | Others | | |
|---|---|---|---|---|---|
| | Learning | | Learning | | Optimization |
| | **Fine** | Coarse | [60] | [37] | [26] |
| Mean | **1.84 mm** | 2.03 mm | 2.19 mm | 2.11 mm | **1.59 mm** |
| SD | **0.38 mm** | 0.52 mm | 0.54 mm | 0.46 mm | **0.30 mm** |
| Time | **4 ms** | **4 ms** | **4 ms** | **4 ms** | 120 s |

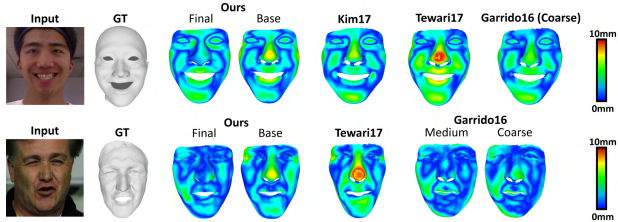

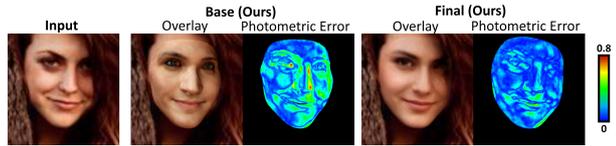

Figure 10. We obtain higher quality than previous learning-based approaches on FaceWarehouse [17] and Volker [64].

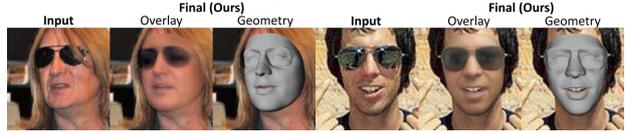

Figure 11. Euclidean photometric error in RGB space, each channel in [0, 1]. Our final results significantly improve fitting quality.

## 7.2. Quantitative Results

We evaluated our approach quantitatively. For geometry, we use the FaceWarehouse [17] dataset and reconstruct 180 meshes (9 identities, 20 expressions each). We compare various approaches, after alignment (rigid transform plus isotropic scaling), to the provided ground truth using the Hausdorff distance. Our approach outperforms the learning-based techniques of Tewari et al. [60] and Kim et al. [37], see Tab. 1. We come close to the high-quality optimization approach of Garrido et al. [26], while being orders of magnitude faster (4ms vs. 120sec) and not requiring feature detection at test time, see Fig. 10 (top). [17] contains mainly 'clean' faces without make-up or beards, since this causes problems even for high-quality offline 3D reconstruction approaches. Our interest is in robustly handling this harder scenario, in which we demonstrate that our approach significantly outperforms previous approaches, see Figs. 6, 8, and 9. We also evaluate our approach on a video sequence (300 frames) with challenging expressions and a characteristic face, which is outside the span of the 3DMM. The ground truth has been obtained by Valgaerts et al. [64]. The results can be found in Tab. 2 and in Fig. 10 (bottom), where it can be seen that our method outperforms other learning- and optimization-based approaches [26, 60]. We evaluate the photometric fitting error of our approach on our validation set, see Fig. 11. Our final results (mean: 0.072, SD: 0.020) have significantly lower error (distance in RGB space, channels in [0, 1]) than the base level (mean: 0.092, SD: 0.025) due to our learned corrective basis.

Table 2. On the Volker sequence, our approach outperforms the results of [26], even if their fixed shape correctives are employed.

| | **Ours** | | Others | | |
|---|---|---|---|---|---|
| | Learning | | Learning | Optimization [26] | |
| | **Fine** | Coarse | [60] | Medium | Coarse |
| Mean | **1.77 mm** | 2.16 mm | 2.94 mm | 1.97 mm | 1.96 mm |
| SD | **0.29 mm** | 0.29 mm | 0.28 mm | 0.41 mm | 0.35 mm |

Figure 12. External occluders are baked into our correctives.

## 8. Limitations

We demonstrated high-quality monocular reconstruction at over 250Hz, even in the presence of facial hair, or for challenging faces. Still, our approach has a few limitations, which can be addressed in future work: External occlusion, e.g., by glasses, are baked into our correctives, see Fig. 12. Resolving this would require a semantic segmentation of the training corpus. We can not guarantee the consistent reconstruction of occluded face regions. We enforce low-dimensionality of our corrective space for robust model learning. Thus, we can not recover fine-scale surface detail. We see this as an orthogonal research direction, which has already produced impressive results [48, 49, 54].

## 9. Conclusion

We have presented the first approach that jointly learns a face model and a parameter regressor for face shape, expression, appearance and illumination. It combines the advantages of 3DMM regularization with the out-of-space generalization of a learned corrective space. This overcomes the disadvantages of current approaches that rely on strong priors, increases generalization and robustness, and leads to high quality reconstructions at over 250Hz. While in this work we have focused on face reconstruction, our approach is not restricted to faces only as it can be generalized to further object classes. As such, we see this as a first important step towards building 3D models from in-the-wild images.

**Acknowledgements:** We thank True-VisionSolutions Pty Ltd for providing the 2D face tracker. We thank E. Richardson, M. Sela, J. Thies and S. Zafeiriou for running their approaches on our images. We thank J. Wang for the video voice-over. This work was supported by the ERC Starting Grant CapReal (335545), the Max Planck Center for Visual Computing and Communication (MPC-VCC), and by Technicolor.

# Self-supervised Multi-level Face Model Learning for Monocular Reconstruction at over 250 Hz
# — Supplemental Material —

Ayush Tewari[1,2]  Michael Zollhöfer[1,2,3]  Pablo Garrido[1,2]  Florian Bernard[1,2]
Hyeongwoo Kim[1,2]  Patrick Pérez[4]  Christian Theobalt[1,2]
[1]MPI Informatics  [2]Saarland Informatics Campus  [3]Stanford University  [4]Technicolor

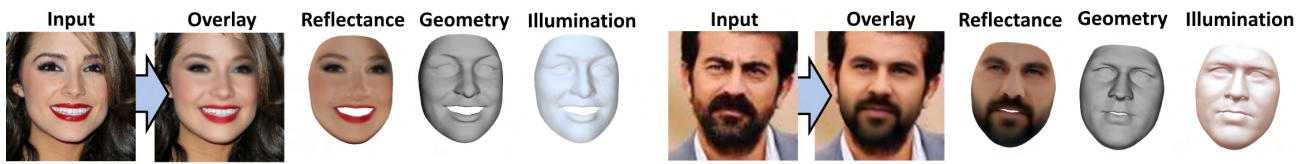

Our novel monocular reconstruction approach estimates high-quality facial geometry, skin reflectance (including facial hair) and incident illumination at over 250 Hz. A trainable multi-level face representation is learned jointly with the feed forward inverse rendering network. End-to-end training is based on a self-supervised loss that does not require dense ground truth.

In this supplemental document we provide more details on our approach. More specifically, we discuss robust training (Sec. 1) of our architecture, we provide the runtime (Sec. 2) for training and testing, we discuss different corrective spaces that we tested (Sec. 3), we perform a study (Sec. 4) on the number of required corrective parameters, and show more results (Sec. 5).

Please note that all shown colored reconstructions of our approach are not textured with the input image, but show the color due to the reconstructed reflectance and/or illumination, for e.g. see Fig. 1. The underlying model is a multi-level face model with a base level employing a 3DMM (Base), and a final level that adds optimal learned per-vertex shape and reflectance correctives (Final). Skin reflectance is represented with a low-dimensional coefficient vector of size $580 = 500 + 80$ (500 coefficients for the correctives and 80 coefficients for the 3DMM). Thus, skin reflectance is stored using only 2.3KB (one float per coefficient). The shape is represented based on a low-dimensional vector of size $644 = 500 + 80 + 64$ (500 coefficients for the correctives, 80 coefficients for the shape identity in the 3DMM, and 64 blendshape coefficients). Thus, the geometry is also stored using only 2.6KB (one float per coefficient). In total, the complete reconstruction is efficiently represented with less than 5KB. This can be exploited for compression, i.e., if the reconstruction has to be transmitted over the internet.

## 1. Training

In the following, we describe how we train our novel architecture end-to-end based on a two stage training strat-egy. Training the face regressor and the corrective space jointly, in one go, turned out to be challenging. For robust training, we first pretrain our network up to the base level for 200k iterations with a learning rate of 0.01. We implemented our approach using *Caffe* [3]. Training is based on *AdaDelta* with a batch size of 5. We use fixed weights $w_\bullet$ in all our experiments. For training the base level we use the following weights $w_{\text{photo}} = 1.9, w_{\text{reg}} = 0.00003, w_{\text{rstd}} = 0.002, w_{\text{smo}} = 0.0, w_{\text{ref}} = 0.0, w_{\text{glo}} = 0.0$ and $w_{\text{sta}} = 0.0$. In addition, we only use the photometric alignment term on the base level. Afterwards, we finetune our complete network for 190k iterations end-to-end with a learning rate of 0.001 for the base level network, 0.005 for the geometry correctives network and 0.01 for the reflectance correctives network. For finetuning, in all our experiments we instantiate our loss based on the following weights $w_{\text{photo}} = 0.2, w_{\text{reg}} = 0.003, w_{\text{rstd}} = 0.002, w_{\text{smo}} = 3.2 \cdot 10^4, w_{\text{ref}} = 13, w_{\text{glo}} = 80, w_{\text{sta}} = 0.08$, and use 500 correctives for both geometry and reflectance. Please note, the illumination estimate for rendering the base and final model is not shared between the two levels, but independently regressed. This is due to the fact that a different illumination estimate might be optimal for the coarse and final reconstruction due to the shape and skin reflectance correctives. During finetuning, all weights associated with the correctives receive a higher learning rate ($\times 100$) than the pretrained layers. We found that this two stage strategy enables robust and efficient training of our architecture.

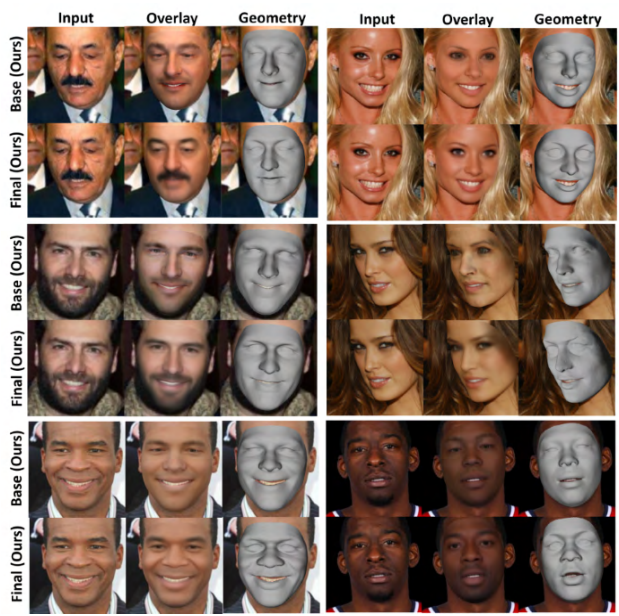

Figure 1. Jointly learning a multi-level model improves the estimated geometry and reflectance compared to the underlying 3DMM on the coarse base level. Note the better aligning nose, lips and the reconstructed facial hair.

## 2. Runtime Performance

We evaluate the runtime performance of our approach on an Nvidia GTX TITAN Xp graphics card. Training our novel monocular face regressor takes 16 hours. A forward pass of our trained convolutional face parameter regressor takes less than 4 ms. Thus, our approach performs monocular face reconstruction at more than 250 Hz.

## 3. Evaluation of the Corrective Space

Our corrective space is based on (potentially non-linear) mappings $\mathcal{F}_\bullet : \mathbb{R}^C \to \mathbb{R}^{3N}$ that map the $C$-dimensional corrective parameter space onto per-vertex corrections in shape or reflectance. The mapping $\mathcal{F}_\bullet(\delta_\bullet | \Theta_\bullet)$ is a function of $\delta_\bullet \in \mathbb{R}^C$ that is parameterized by $\Theta_\bullet$. In the linear case, one can interpret $\Theta_\bullet$ as a matrix that spans a subspace of the variations, and $\delta_\bullet$ is the coefficient vector that reconstructs a given sample using the basis. Let $\mathcal{L}_i(\delta) = \mathbf{M}_i \delta + \mathbf{b}_i$ be an affine/linear mapping and $\Theta_\bullet^{[i]}$ stack all trainable parameters, i.e., the trainable matrix $\mathbf{M}_i$ and the trainable offset vector $\mathbf{b}_i$. We tested different linear and non-linear corrective spaces, see Fig. 2. The affine/linear model (Linear) is given as:

$$\mathcal{F}_\bullet(\delta_\bullet | \Theta_\bullet^{[0]}) = \mathcal{L}_0(\delta_\bullet) \ . \quad (1)$$

However, in general we do not assume $\mathcal{F}_\bullet$ to be affine/linear. For this evaluation, given the non-linear function $\Psi$, which in our case is a ReLU non-linearity, we define a corrective model with two affine/linear layers and one

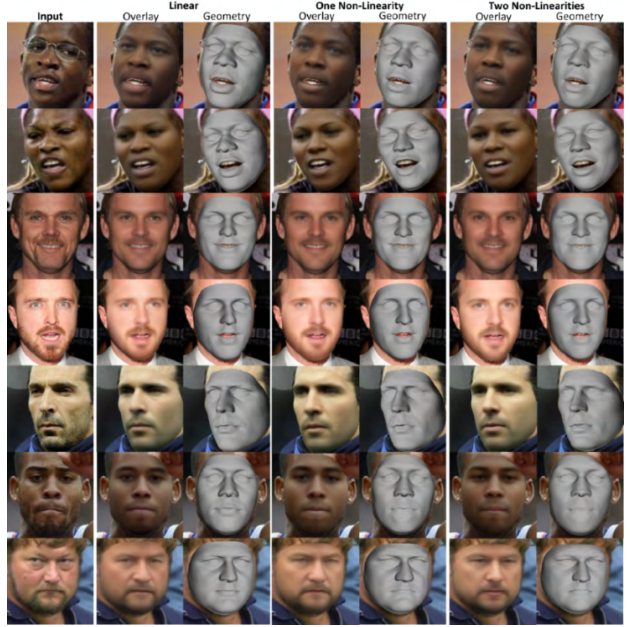

Figure 2. Comparison of linear and non-linear corrective spaces.

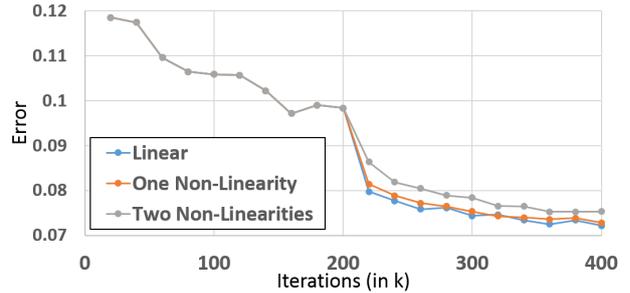

Figure 3. The affine/linear model obtains a lower photometric re-rendering error than the two tested non-linear corrective spaces. Thus, we use the affine/linear model in all our experiments.

non-linearity (One Non-Linearity):

$$\mathcal{F}_\bullet(\delta_\bullet | \Theta_\bullet^{[0]}, \Theta_\bullet^{[1]}) = \mathcal{L}_1(\Psi(\mathcal{L}_0(\delta_\bullet))) \ . \quad (2)$$

In addition, we also define a corrective model with three affine/linear layers and two non-linearities (Two Non-Linearities):

$$\mathcal{F}_\bullet(\delta_\bullet | \Theta_\bullet^{[0]}, \Theta_\bullet^{[1]}, \Theta_\bullet^{[2]}) = \mathcal{L}_2(\Psi(\mathcal{L}_1(\Psi(\mathcal{L}_0(\delta_\bullet))))) \ . \quad (3)$$

As can be seen in Fig. 2, the results obtained by the affine/linear and non-linear models are visually quite similar. The affine/linear model obtains a lower photometric re-rendering error, see Fig. 3. Thus, we decided for the simpler affine/linear model and use it in all our experiments.

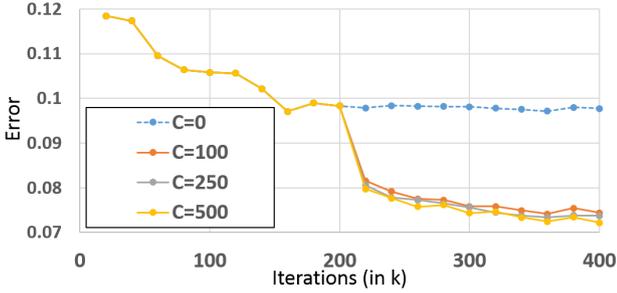

Figure 4. We trained our network with $C = 0, 100, 250$ and $500$ corrective parameters for shape and reflectance. Our networks with correctives significantly improve upon the baseline network that only uses the 3DMM ($C = 0$) in terms of the photometric re-rendering error. The corrective basis with $C = 500$ parameters achieves the lowest photometric re-rendering error. Thus, we use this network for all further experiments.

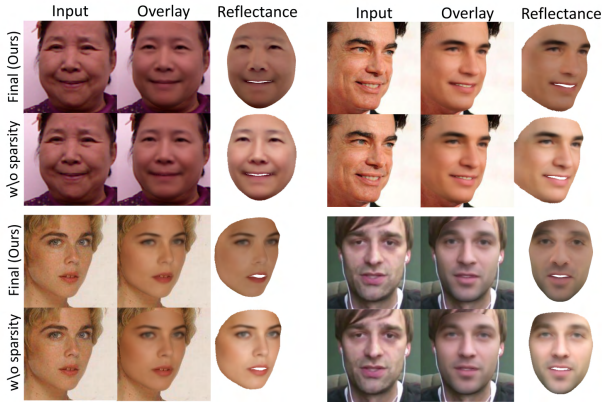

Figure 5. Removing reflectance sparsity leads to shading information being wrongly explained by reflectance variation.

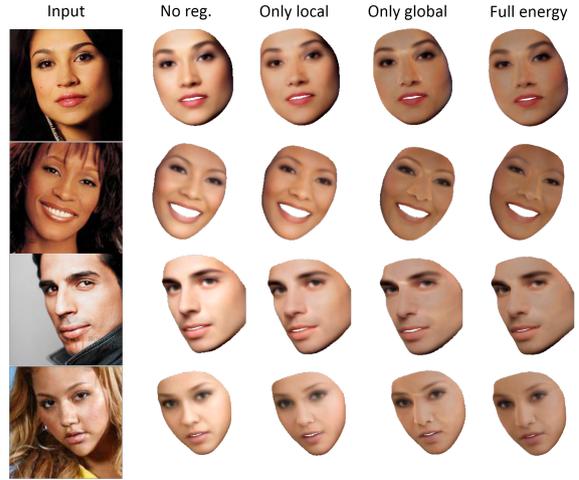

Figure 6. Both local sparsity and global constancy terms help in obtaining plausible reflectance estimates.

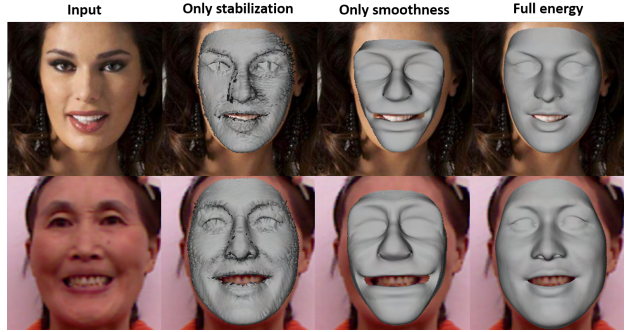

Figure 7. Both corrective smoothness and stabilization terms help in obtaining nice geometry estimates.

## 4. Additional Evaluation

We also report the photometric re-rendering error on a test set (2k images) for a varying number of corrective parameters. To this end, we follow our two-stage training schedule and trained our network with $C = 0, 100, 250$ and $500$ corrective parameters for shape and reflectance, see Fig. 4. Our networks with correctives significantly improve upon the baseline network that only uses the 3DMM ($C = 0$) in terms of the photometric re-rendering error. The best results in terms of the lowest photometric re-rendering error are obtained by our network with $C = 500$ additional corrective parameters for shape and reflectance. Thus, we use this network for all further experiments.

We also performed an ablation study to evaluate the contribution on reconstruction quality of the different objectives of our self-supervised loss function. More specifically, we evaluated two different variations of our self-supervised loss: 1) We removed all corrective shape regularizers and 2) We removed all reflectance sparsity priors. Removing the shape regularizers (stabilization and smoothness) leads to a complete failure during training, since all corrective per-vertex displacements are independent and thus the reconstruction problem is severely underconstrained. If the reflectance sparsity priors (local and global) are removed, the network can still be trained and the overlayed reconstructions look plausible, see Fig. 5, but all shading information is wrongly explained by reflectance variation. Thus, both the used shape and reflectance priors are necessary and drastically improve reconstruction quality. An ablative analysis of the individual terms of the reflectance and shape regularizers can be found in Figs. 6 and 7. We also evaluate our estimated reflectance on synthetic images. We generate a synthetic training corpus using 80 reflectance vectors of the 3DMM. We only allow the base model of the network to regress 40 reflectance parameters, which allows the final model to learn the reflectance correctives between the base model and the ground truth, see Fig. 8. We also perform a quantitative evaluation by computing per-pixel RGB distances between the rendered reflectance image and the ground truth (we compensate for global shifts in reflectance). We render these images using the mean face geometry in a canonical pose. We obtain an error of 0.072 (averaged over 1k test images), which shows the accuracy of our predictions.

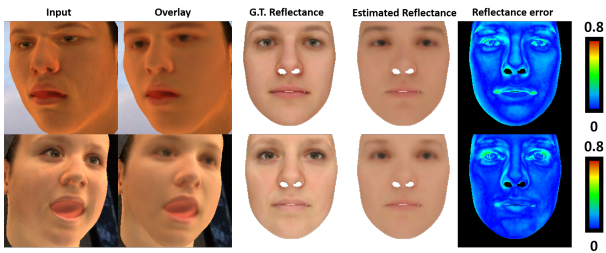

Figure 8. Our reflectance estimates are close to the ground truth for synthetic images.

## 5. Additional Results and Comparisons

We show more results (Fig. 13) and comparisons to current optimization-based (Figs. 10 and 11) and learning-based (Figs. 14 and 15) state-of-the-art approaches. Note, in the comparison to Tewari et al. [7], we compare to their weakly supervised training, which, similar to our approach, uses a sparse set of facial landmarks for supervision. Our approach obtains high reconstruction quality and compares favorably to all of these state-of-the-art techniques. In particular, it is able to reconstruct colored surface reflectance and it robustly handles even challenging cases, such as occlusions by facial hair and make-up. For a detailed discussion of the differences to the other approaches we refer to the main document. In addition, we show more examples of limitations (Fig. 12), such as external occluders, which are baked into the recovered model. We also show more examples of the photometric re-rendering error (Fig. 9) and show that the corrective space improves the regressed shape and reflectance estimate (Fig. 1).

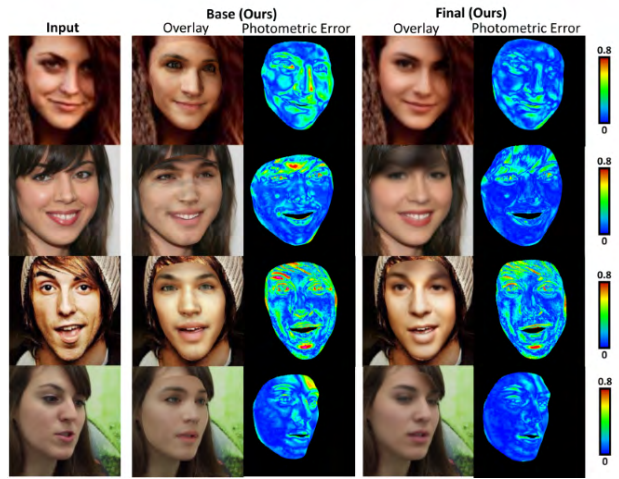

Figure 9. Euclidean photometric error in RGB space, each channel in [0, 1]. Our final results significantly improve fitting quality.

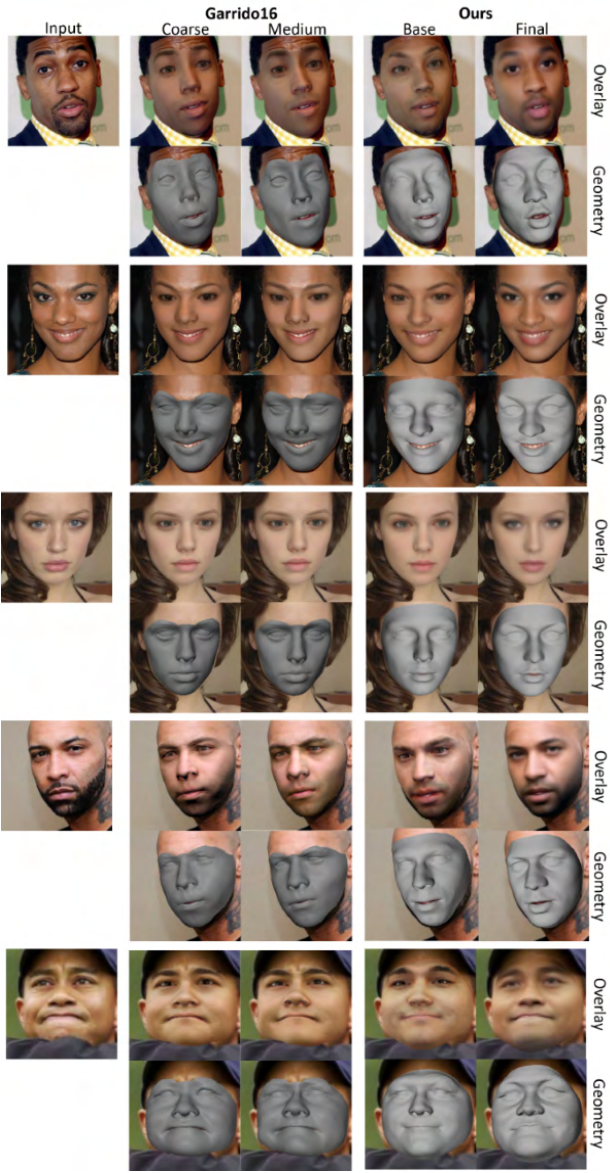

Figure 10. Comparison to Garrido et al. [2]. We achieve higher quality reconstructions, since our jointly learned model allows leaving the restricted 3DMM subspace and generalizes better than a corrective space based on manifold harmonics.

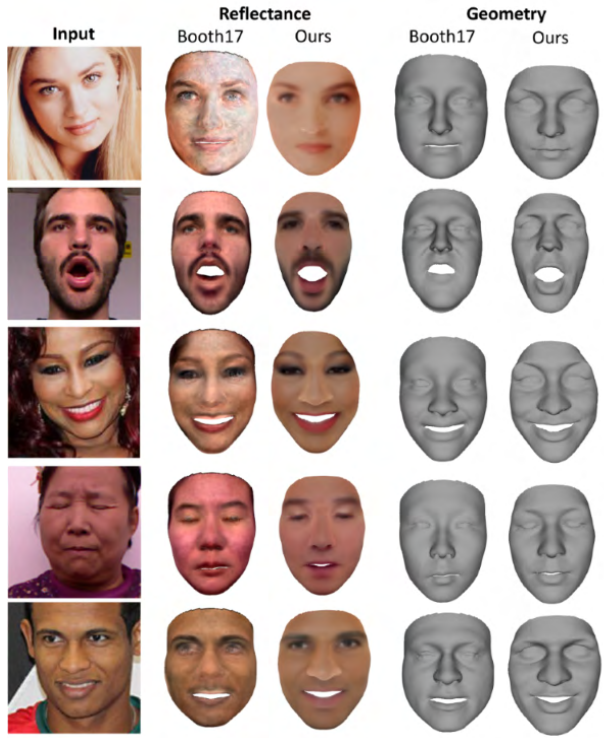

Figure 11. In contrast to the 'in-the-wild' texture model of Booth et al. [1] that contains shading, our approach yields a reflectance model. In addition, our learned optimal corrective space goes far beyond the restricted low-dimensional geometry subspace that is commonly employed.

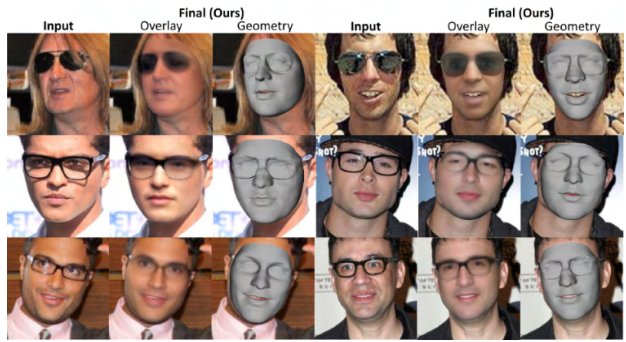

Figure 12. External occluders might be baked into the correctives.

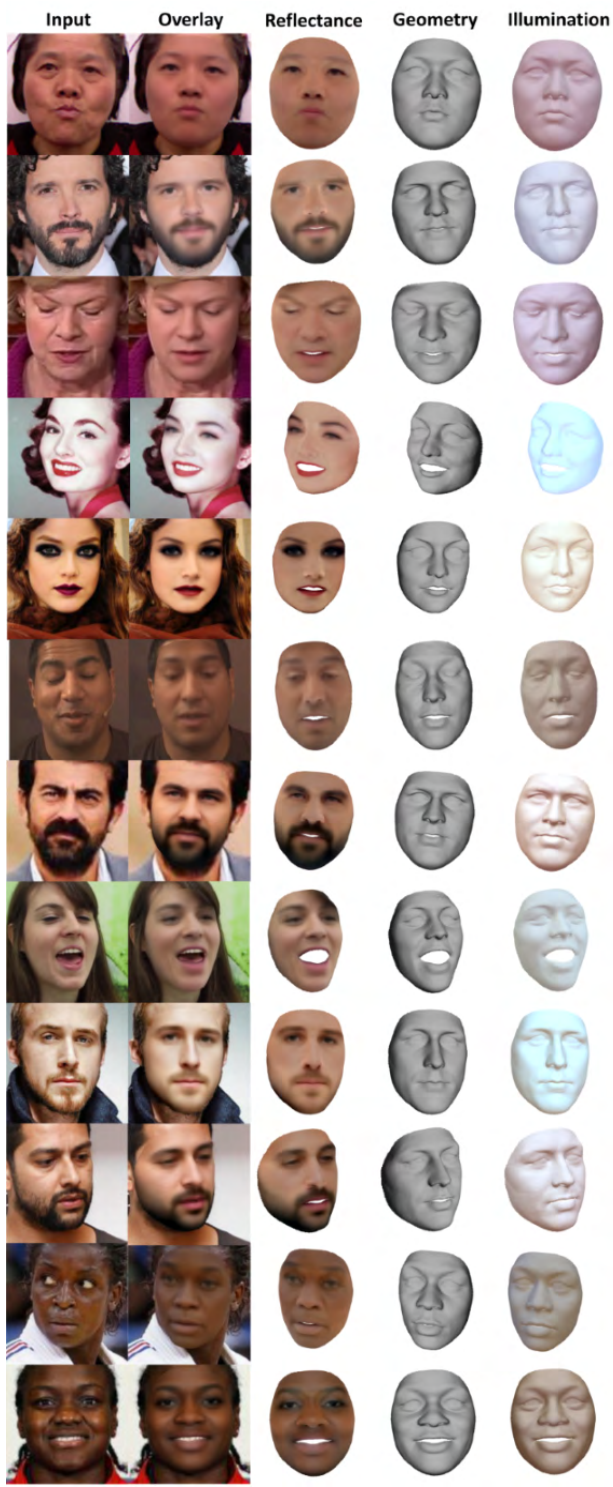

Figure 13. Our approach allows for high-quality reconstruction of facial geometry, reflectance and incident illumination from just a single monocular color image. Note the reconstructed facial hair, e.g., the beard, reconstructed make-up, and the eye lid closure, which are outside the restricted 3DMM subspace.

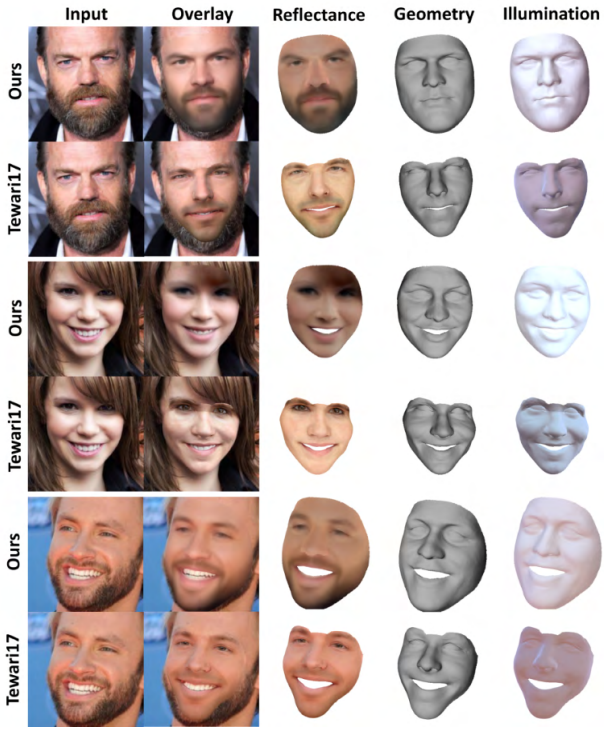

Figure 14. Comparison to Tewari et al. [7]. We achieve higher quality in terms of geometry and reflectance, since our jointly trained model allows leaving the restricted 3DMM subspace. This prevents surface shrinkage due to unexplained facial hair.

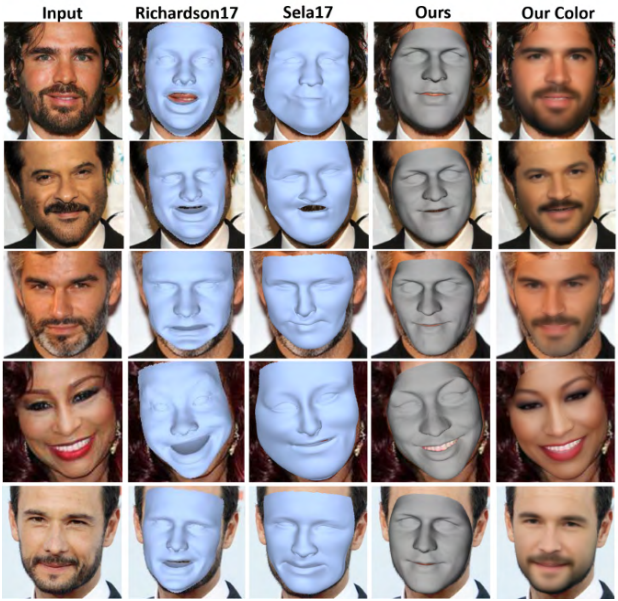

Figure 15. Comparison to Richardson et al. [4, 5] and Sela et al. [6]. They obtain impressive results for faces within the span of the synthetic training corpus, but suffer for out-of-subspace shape and reflectance variations, e.g., people with beards. Our approach is not only robust to facial hair and make-up, but learns to reconstruct such variations based on the jointly trained model.